# An Unsupervised Learning Classifier with Competitive Error Performance

Daniel N. Nissani (Nissensohn)
dnissani@post.bgu.ac.il

**Abstract.** An unsupervised learning classification model is described. It achieves classification error probability competitive with that of popular supervised learning classifiers such as SVM or kNN. The model is based on the incremental execution of small step shift and rotation operations upon selected discriminative hyperplanes at the arrival of input samples. When applied, in conjunction with a selected feature extractor, to a subset of the ImageNet dataset benchmark, it yields 6.2 % Top 3 probability of error; this exceeds by merely about 2 % the result achieved by (supervised) k-Nearest Neighbor, both using same feature extractor. This result may also be contrasted with popular unsupervised learning schemes such as k-Means which is shown to be practically useless on same dataset.

**Keywords**: Unsupervised Learning, Linear Classifiers, Neural Models.

## 1 Introduction

Unsupervised learning methods have been explored significantly during the last decades. Part of this work has been oriented towards the solution of the so called Classification function. Most of this research resulted in schemes which are based on either similarity (or alternatively distance) measures (such as the popular k-Means method and its variants, e.g. MacQueen [9]) or on the estimation of density mixtures (e.g. Duda and Hart [1], Bishop [5]); both approaches require the user to supply significant prior information concerning the underlying model (such as number k of distinct classes for both k-Means and parametric models, and the assumed probability distribution function class for the later).

Our work is inspired by the 'cluster assumption': that real world classes are separated by low probability density regions in an appropriate feature space. This assumption has naturally lead us and others to the so called 'valley seeking' methods which have been explored since the 70's (Lewis et al., [11]) and up to the more recent work by Pavlidis et al. [12]. Many of these works, including this last quoted, require the conduction of a density estimation process (parametric or non-parametric); none of them makes *implicit* use of the density mixture as in our present work; and none of them has provided, to the best of our knowledge, satisfactory experimental results on real life challenging problems such as ImageNet dataset classification. It should be noted that these unsupervised learning schemes, relevant to the aforementioned Classification function, should be distinguished from, and should not be confused with, unsupervised learning schemes relevant to the canonically associated Feature Extraction function, such as Generative Adversarial Networks (GAN, e.g. Donahue et al., [2]), Variational Autoencoders (VAE, e.g. Pu et al. [3]), Restricted Boltzmann Machines (RBM, e.g. Ranzato and Hinton [4]), Transfer Learning (Yosinski et al. [10]), etc.

In the sequel we present initial results on an unsupervised learning classifier exhibiting error performance competitive to that of popular near optimal supervised learning classifiers such as k-Nearest Neighbors (kNN) and Support Vector Machine (SVM). The proposed model is presented in the next Section. In Section 3 we present simulation results; in Section 4 we summarize our work and make a few concluding remarks.

## 2 Proposed Model

Our model deals with the unsupervised learning Classifier function of a Pattern Recognition machine. Measurement vectors $\mathbf{v} \in R^D$, where D is the dimension of the measurement or observed space, are fed into a Feature Extractor. These vectors may consist of image pixel values, digitized speech samples, etc. These vectors **v** are mapped by the Feature Extractor into so called feature vectors $\mathbf{x} \in R^d$, where d denotes the dimension of the feature space. Such a transformation or mapping is supposedly capable of yielding better inter-class separability properties. The feature vectors **x** are then fed into a Classifier whose ultimate role is to provide an output code vector **y** (or equivalently a label, or a class name) providing discrete information regarding the class to which the input measurement vector **v** belongs. The classifier may be of supervised learning type in which case pre-defined correct labels are fed into it simultaneously with input vectors **v** during the so called training stage, or of unsupervised learning type, as in our case, where no such labels are presented at any time.

It is a fundamental assumption of our model that the feature space vectors are distributed in accordance with some unknown probability distribution mixture. This mixture is composed of the weighted sum of conditional distribution densities. We assume that the bulks of this mixture occupy a bounded region of feature space. We further assume that said mixture obeys a sufficient condition, informally and loosely herein stated, namely that the intersecting slopes of the weighted conditionals form well defined low density regions, or 'valleys' (the so called 'cluster assumption'). Conditional probability densities should be normalizable by definition; since this forces their density to eventually decay (or vanish) then this sufficient condition is reasonable and commonly met. Note that in spite this assumption our model is not parametric, that is no assumption is made regarding the functional form of the distributions involved, nor about the number of presented classes. The resultant valleys can, in general, take the form of non-linear hyper-surfaces; like any linear classifier (e.g. linear SVM) we finally assume that, by virtue of an appropriate Feature Extraction function, these hyper-surfaces can be reasonably approximated by simple hyperplanes.

Our proposed model is initiated by populating the relevant feature space bounded region, with a certain number, or pool, of hyperplanes. These can be either uniformly randomly, or orderly and uniformly grid-wise distributed within this region. Following this initiation, input feature vectors are presented. Upon arrival of one such input vector, each such hyperplane generates an output signal which indicates whether this input sample resides on one side of the hyperplane or the other. In addition, in response to this input vector, some of the hyperplanes are shifted and/or rotated by small increments, gradually migrating from regions characterized by high probability density to regions of low probability density, and thus converging (in the mean) towards said density mixture 'valleys', by means of a mechanism to be described in the sequel.

It is convenient to commence description of our model by means of a simpler model; refer to Figure 1 which depicts a 1-dimensional probability density mixture $p(\mathbf{x})$ along with some fundamental entities related to our proposed model. When conditional densities $p(\mathbf{x}|C_i)$ of this mixture are sufficiently spaced and have appropriate priors $p(C_i)$ their mixture creates a minimal point (a 'valley') such as point **a** in Figure 1. The 2-classes mixture density of Figure 1 can be expressed as

$$p(\mathbf{x}) = p(C_1)\, p(\mathbf{x}|C_1) + p(C_2)\, p(\mathbf{x}|C_2) \tag{1}$$

In our simple model of Figure 1 there exists a single hyperplane (reduced to a single point in our 1-dimensional model) defined by the hyperplanar semi-linear operator (other hyperplanar operators may be considered)

$$y_{t+1} = \max\{0,\ \mathbf{w}_t \,.\, \mathbf{x} - \theta_t\} \tag{2}$$

where the hyperplane weight vector is $\mathbf{w}_t \in R^d$ (d = 1, and $\mathbf{w}_t = 1$ is assumed in this introductory model), the hyperplane threshold variable is $\theta_t \in R$, the t subscript indicates the hyperplane self-time, "." denotes inner product operation, and where individual hyperplane indexes (only one in this model) are omitted for readability.

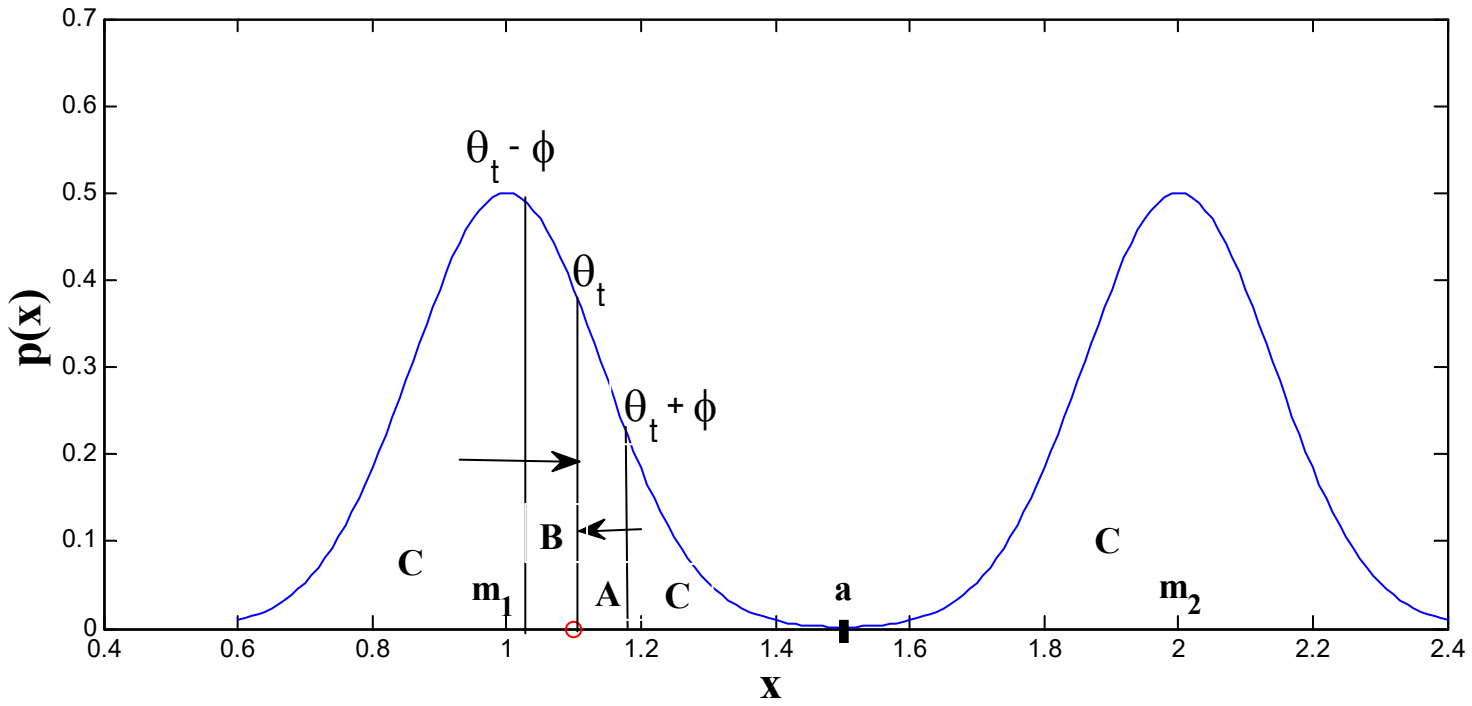

**Figure 1**: A 1-dimensional density mixture and proposed model related entities

Hyperplanes hence act in this model as linear discriminator functions. One of these hyperplanes is indicated in Figure 1 by means of its associated hyperplane threshold variable $\theta_t$. The present state of this variable, in between the distribution mixture two peaks or modes, ensures (with high probability) the model arrival to stable equilibrium, as will momentarily become evident.

We now propose the following simple Shift unsupervised learning rule, applied upon arrival of each input vector **x**:

$$\theta_{t+1} = \begin{cases} \theta_t - \varepsilon, & \theta_t + \Phi \geq \mathbf{w}_t \, . \, \mathbf{x} > \theta_t \\ \theta_t + \varepsilon, & \theta_t - \Phi \leq \mathbf{w}_t \, . \, \mathbf{x} \leq \theta_t \\ \theta_t \quad , & \text{otherwise} \end{cases} \tag{3}$$

where $\Phi$ and $\varepsilon$ are two small positive scalar model parameters. In accordance with this rule a new feature vector **x** may either: fall in area **A** in which case $\theta_t$ will be Shifted to the left; fall in area **B** which will Shift $\theta_t$ to the right; or fall in any of areas **C** which will maintain $\theta_t$ unchanged. Since events of class **B** have greater probability than events of class **A** (area of region **B** is greater than area of region **A**) there will be a net 'pressure' to Shift $\theta_t$ to the right as indicated by the arrows of different size in Figure 1. Such Shift operations will tend to carry $\theta_t$ from regions of high probability density to regions of low probability density. Once $\theta_t$ arrives to the vicinity of the distribution 'valley' (at about $\theta_t$ = **a**) stochastic stable equilibrium will be achieved: equal pressures will be exerted on both directions.

We note that increasing the Shift step $\varepsilon$ may accelerate convergence, but may also increase the probability of jumping over the 'top of the hill' to the left of the leftmost mode, thereafter drifting towards a non-discriminatory and less useful direction. Similarly, increasing $\Phi$ may also accelerate convergence (since it reduces probability of inactive C events) but, again, may disrupt convergence direction. Both $\Phi$ and $\varepsilon$ should be (and were) fine tuned during simulation. Proof of convergence (in the mean) of $\theta_t$ to the point **a** is simple, under some idealized assumptions, and is omitted herein. Note that both $\mathbf{m_1}$ and $\mathbf{m_2}$ mixture peaks are also stochastic equilibrium points, though unstable, as can be also easily proved.

The model initial condition $\theta_0$ affects, along with the values of $\varepsilon$ and $\Phi$ parameters, the probability of converging to a useful discrimination point like **a**: the more $\theta_0$ resides away of any peak (but in between peaks) the larger the probability of such a useful convergence; and vice versa.

As mentioned above our herein proposed model achieves competitive performance in the sense of classification error probability. Its error performance is close to that of an optimal supervised learning classifier in as much as the mixture density valley is close (in some suitable measure) to the optimal hyper-surface, and in as much as the final state of the model hyperplane deviates from this valley (due to stochastic fluctuations, finite valued parameters effects, etc.). This optimal hyper-surface in turn depends on the mixture density conditionals and priors and is the solution to the equation $p(C_1)\ p(\mathbf{x}|C_1) - p(C_2)\ p(\mathbf{x}|C_2) \equiv f(\mathbf{x}) = 0$. If both classes have symmetric identical conditionals (except of course their peaks position) and have equal priors, then the mixture optimal discriminant classifier lies, by symmetry, at $\theta_t$ = **a**, midway between the two peaks $\mathbf{m_1}$ and $\mathbf{m_2}$. Our proposed method on the other hand, converges in this case (in the mean and neglecting the fore-mentioned deviations) to a minima of p(**x**), and is thus the zero derivative solution of Equation (1) (i.e. p'(**x**) = 0) which results, in said symmetric case, in exactly the same point **a** and thus achieves (in the mean, and neglecting said deviations) optimality. If priors and/or conditionals differ from each other, both solutions (of f(**x**) = 0 and of p'(**x**) = 0) will slowly drift away from each other, and only near-optimality is achieved. The possible distance between the valley and the optimal hyper-surface may be considered the fundamental penalty paid, in lack of labels, for our limited ability to estimate priors and conditionals.

We now turn to revise and modify our model to adapt it to high dimensional feature spaces. The above considerations may be directly extended to these spaces, wherein the valley single point becomes a hyperplane. We will see that all is needed is the addition of a Rotation operation to the fore-mentioned Shift operation, which was described above.

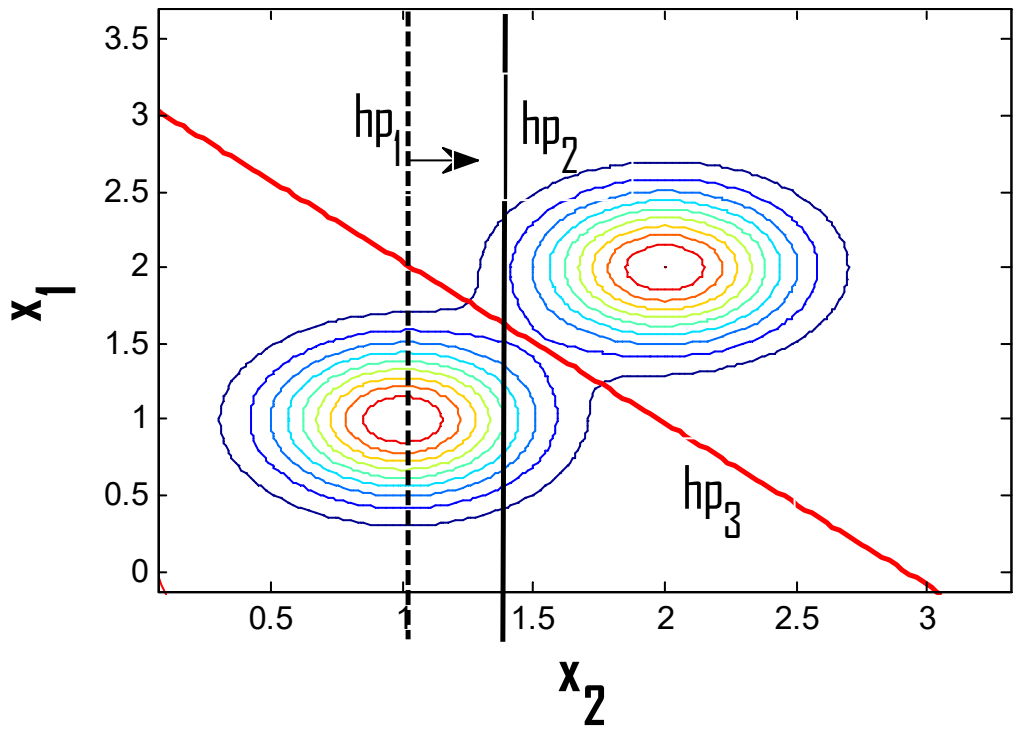

**Figure 2**: 2-dimensional feature space – the need for a Rotation Learning Rule

To see why Shift operation alone is not enough refer to Figure 2 which shows 2 unimodal classes, described by means of their probability iso-density contours in a 2-dimensional space. The discriminating hyperplane initial condition is assumed to be $\mathbf{hp_1}$ and the 'valley' near-optimal hyperplane position is $\mathbf{hp_3}$. It is clear that a sequence of Shift only operations can lead discrimination to an equilibrium point hyperplane like $\mathbf{hp_2}$, but is not capable of converging to the near optimal $\mathbf{hp_3}$. The addition of a Rotation operation is apparently required.

Refer now to Figure 3 which shows a 2-dimensional space containing 2 distinct classes schematically represented by their means $\mu_1$ and $\mu_2$. These means reside relatively close to and on both sides of a hyperplane $\mathbf{hp_1}$. Figure 3 presents also an assumed near optimally placed hyperplane $\mathbf{hp_2}$, and an input vector **x** also relatively close to hyperplane $\mathbf{hp_1}$. The hyperplane $\mathbf{hp_1}$ is defined by those vectors **x** which satisfy the equation $\mathbf{w} \cdot \mathbf{x} = \theta$ where we have omitted time subscripts for clarity, and where, by convention, the weight vector **w** is normalized (i.e. $\|\mathbf{w}\| = 1$). As is well known, **w** defines the hyperplane $\mathbf{hp_1}$ orientation relative to the feature space axes and is orthogonal to this hyperplane, while θ defines the hyperplane distance from the axes origin. It is similarly well known that rotation in high dimensional spaces is defined by a 2-plane of rotation, by a rotation point, and by a rotation angle, and that it keeps invariant the (n-2)-subspace orthogonal to this 2-plane (this invariant subspace reduces, when we deal with 3-dimensional spaces, to the familiar 'axis of rotation', a meaningless term in high dimensional spaces).

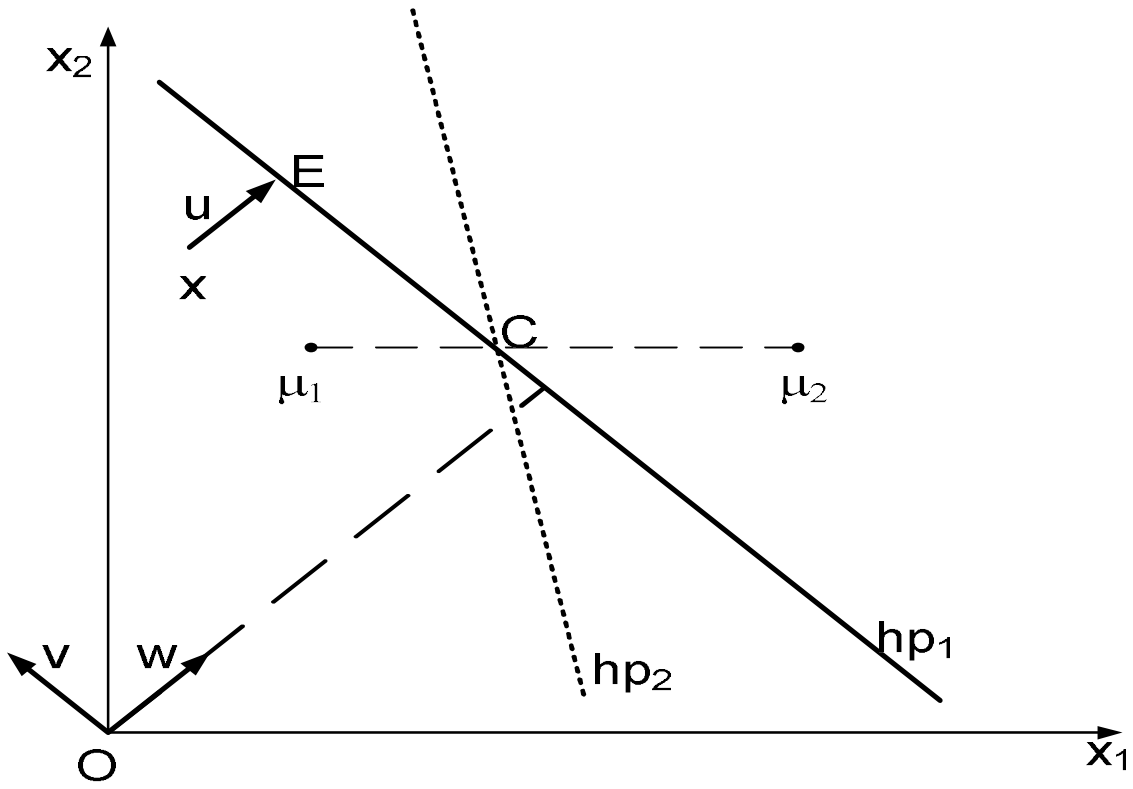

**Figure 3**: Entities related to the Rotation Learning Rule

We finally denote by **C** the intersection point between $\mathbf{hp_1}$ and the segment connecting $\mu_1$ and $\mu_2$, and by **E** the orthogonal projection of **x** upon $\mathbf{hp_1}$. It is proposed that, upon arrival of some feature vector **x**, within $\Phi$ distance from $\mathbf{hp_1}$, we rotate $\mathbf{hp_1}$ by a small angle $\alpha \ll 1$, around the point **C**, over a rotation 2-plane defined by the normalized orthogonal vector **u** from **x** to $\mathbf{hp_1}$ (which equals, up to possible sign inversion, to **w**) and by the normalized vector **v** connecting **E** and **C;** to gain better intuition on this proposal the reader is encouraged to visualize the above defined entities within a 3-dimensional space and corresponding probability density mixture. The model contains, as mentioned above, a pool of hyperplanes. It is convenient to assign to each hyperplane a separate self-timer t (t subscripts omitted above for readability) which advances whenever an incoming feature vector **x** falls within $\Phi$ distance from said hyperplane. Consolidating, we propose the following unsupervised learning scheme, executed for each and every hyperplane, upon arrival of a feature vector **x**.

**<u>Shift Operation:</u>**

Exactly as with our 1-dimensional model, repeated herein for readability:

$$\theta_{t+1} = \begin{cases} \theta_t - \varepsilon, & \theta_t + \Phi \geq \mathbf{w}_t \,.\, \mathbf{x} > \theta_t \\ \theta_t + \varepsilon, & \theta_t - \Phi \leq \mathbf{w}_t \,.\, \mathbf{x} \leq \theta_t \\ \theta_t \;, & \text{otherwise} \end{cases} \qquad (4)$$

**Near Classes Means Estimate:**

In real applications the feature space may be populated by many distinct classes. As described above a means estimate of the classes with means close to a hyperplane is required for the Rotation operation of said hyperplane. Such means estimates can be implemented as weighted averages of input vectors, calculated in separate for both sides of each hyperplane; we allow for the weights to be a function of the distance of **x** from the hyperplane, assigning in general a smaller weight to input vectors farther away from it; this allows to disregard distant inputs, possibly belonging to other, faraway, classes. For each and every hyperplane we update $c^1_{t+1}$ and $\hat{\mu}^1_{t+1}$ if $\mathbf{w}_t \,.\, \mathbf{x} \leq \theta_t$ and update $c^2_{t+1}$ and $\hat{\mu}^2_{t+1}$ otherwise:

$$c^j_{t+1} = c^j_t + g(|\mathbf{w}_t \,.\, \mathbf{x} - \theta_t|\,;\beta), \qquad c^j_0 = 0, \tag{5}$$

and

$$\hat{\mu}^j_{t+1} = (c^j_t\, \hat{\mu}^j_t + g(|\mathbf{w}_t \,.\, \mathbf{x} - \theta_t|\,;\beta)\,\mathbf{x}) \,/\, c^j_{t+1}\,, \tag{6}$$

where $\hat{\mu}^j_t$ denote mean estimates, j = 1 or 2 denote each of both hyperplane half spaces $\mathbf{w}_t \,.\, \mathbf{x} \leq \theta_t$ or $\mathbf{w}_t \,.\, \mathbf{x} > \theta_t$ respectively, $c^j_t$ are the cumulative weights associated with this hyperplane, and where $g(|\,.\,|;\ \beta)$ is a distance dependent weight function with parameter $\beta$.

A typical and simple weight function could consist of a uniform positive weight for vectors **x** satisfying $\theta_t - \beta \leq \mathbf{w}_t \,.\, \mathbf{x} \leq \theta_t + \beta$ with some $\beta > 0$, and zero weight otherwise. We assume the use of this simple weight function in the sequel. To initiate these mean estimates we may simply use the first incoming sample (within appropriate distance from the hyperplane) for one side estimate and a symmetrically reflected virtual point for the other, which results, in the case of the uniform weight function suggested above, in

$$\hat{\mu}^1_0 = \mathbf{x}, \quad \hat{\mu}^2_0 = \mathbf{x} + 2\,|\,\mathbf{w}_t \,.\, \mathbf{x} - \theta_t\,|\,\mathbf{w}_t\,, \qquad \text{if } \theta_t - \beta \leq \mathbf{w}_t \,.\, \mathbf{x} \leq \theta_t \tag{7}$$

$$\hat{\mu}^2_0 = \mathbf{x}, \quad \hat{\mu}^1_0 = \mathbf{x} - 2\,|\,\mathbf{w}_t \,.\, \mathbf{x} - \theta_t\,|\,\mathbf{w}_t\,, \qquad \text{if } \theta_t < \mathbf{w}_t \,.\, \mathbf{x} \leq \theta_t + \beta \tag{8}$$

This initiation scheme allows allocating to $\hat{\mu}^2_0$ (or alternatively to $\hat{\mu}^1_0$) some value even though there might exist no class in the neighborhood of the other side of this hyperplane.

**C, E, u and v calculation:**

As mentioned above **C**, **E**, **u** and **v** are also required for the implementation of the Rotation operation. Upon arrival of a feature vector **x**, said calculations are executed only for those hyperplanes for which $\theta_t + \Phi \geq \mathbf{w}_t \,.\, \mathbf{x} \geq \theta_t - \Phi$. Inspection of Figure 3 and simple vector algebra manipulations result in:

$$\mathbf{C} = \hat{\mu}^1{}_t + (\theta_t - \mathbf{w}_t \,.\, \hat{\mu}^1{}_t) \,.\, (\hat{\mu}^2{}_t - \hat{\mu}^1{}_t) \,/\, (\mathbf{w}_t \,.\, (\hat{\mu}^2{}_t - \hat{\mu}^1{}_t)) \quad (9)$$

$$\mathbf{E} = \mathbf{x} + (\theta_t - \mathbf{w}_t \,.\, \mathbf{x}\,)\, \mathbf{w}_t \,/\, \|\, \mathbf{w}_t\| \quad (10)$$

$$\mathbf{u} = (\mathbf{E} - \mathbf{x}) \,/\, \|\mathbf{E} - \mathbf{x}\| = \mathrm{sign}(\, \mathbf{w}_t \,.\, \mathbf{x} - \theta_t\,) \,.\, \mathbf{w}_t \,/\, \|\, \mathbf{w}_t\| \quad (11)$$

where sign(.) denotes the signum function (time subscripts omitted from **C**, **E**, **u** and **v**); note the possible sign disagreement between **u** and $\mathbf{w}_t$; and

$$\mathbf{v} = (\mathbf{E} - \mathbf{C}) \,/\, \|\mathbf{E} - \mathbf{C}\| \quad (12)$$

**Rotation Operation:**

Here too, this operation is executed only for those hyperplanes for which $\theta_t + \Phi \geq \mathbf{w}_t \,.\, \mathbf{x} \geq \theta_t - \Phi$. For d-dimensional Rotation formulation we follow the neat (and slightly abusive) vector notation of Teoh [6] (the reader is referred there for details):

$$\mathbf{w}_{t+1} = \mathrm{rot}(\mathbf{w}_t)_{P,\alpha,C} =$$

$$= \mathbf{w}_t + [\mathbf{u} \quad \mathbf{v}] \begin{pmatrix} \cos\alpha - 1 & -\sin\alpha \\ \sin\alpha & \cos\alpha - 1 \end{pmatrix} \begin{pmatrix} (\mathbf{w}_t \,.\, \mathbf{u}) \\ (\mathbf{w}_t \,.\, \mathbf{v}) \end{pmatrix} \quad (13)$$

where P is the rotation 2-plane as defined by the vectors **v** and **u**, **C** is the rotation point (see Figure 3), and $\alpha << 1$ is a small rotation angle for which we may simplify to $\sin\alpha \approx \alpha$ and $\cos\alpha \approx (1 - \alpha^2/2)$. The sense of rotation (clockwise or counter-clockwise) is set such that the distance between the sample point **x** and the rotated hyperplane increases. Similarly to the Shift operation this ensures hyperplane migration (in the mean) toward a lower probability density region as desired.

Rotation also shifts the hyperplane (i.e. changes its distance from O) so it affects its $\theta_t$ variable, and this has to be updated too (in addition to the Shift learning rule update, Equation (4) above); recalling that the point **C** is invariant under Rotation and that it belongs to the hyperplane (rotated or not) then

$$\theta_{t+1} = \mathbf{w}_{t+1} \,.\, \mathbf{C} \quad (14)$$

**Hyperplane Output Code:**

The hyperplane output code may take the form (amongst many other options) of Equation (1) herein repeated for convenience:

$$y_{t+1} = \max\{0,\ \mathbf{w}_t \,.\, \mathbf{x} - \theta_t\} \quad (15)$$

The collection of all scalar outputs $y_{t+1}$ make up together the model output code vector $\mathbf{y}_{t+1} \in R^N$ where N is the size of the hyperplane pool.

**Self Timer Update:**

As mentioned above a self-timer is conveniently assigned to each hyperplane. This self-timer is incremented whenever an input feature vector **x** arrives within distance Φ from said hyperplane, namely (hyperplane index omitted)

$$t \rightarrow \begin{cases} t+1 \quad , \quad \theta_t + \Phi \geq \mathbf{w}_t \,.\, \mathbf{x} \geq \theta_t - \Phi \\ t \quad , \quad \text{otherwise} \end{cases} \tag{16}$$

Just as in our earlier 1-dimensional example, the Shift and Rotation operations will tend (in the mean) to migrate the hyperplanes from regions of high probability density to regions of low probability density. It should be noted that all learning operations (Equations (4) to (14) above) executed upon a hyperplane do exclusively depend on variables of this said hyperplane, and of no other; thus, the proposed learning rules are local, and 'Hebbian' in this sense. The Rotate operation related calculations (Equations (5) to (14) above) for each hyperplane may be advantageously initiated only after the execution of some pre-defined quantity of executed Shift operations for this hyperplane: this allows each hyperplane to land closer to desired (low density) regions in a first stage, making these later calculations more relevant and the convergence process more efficient.

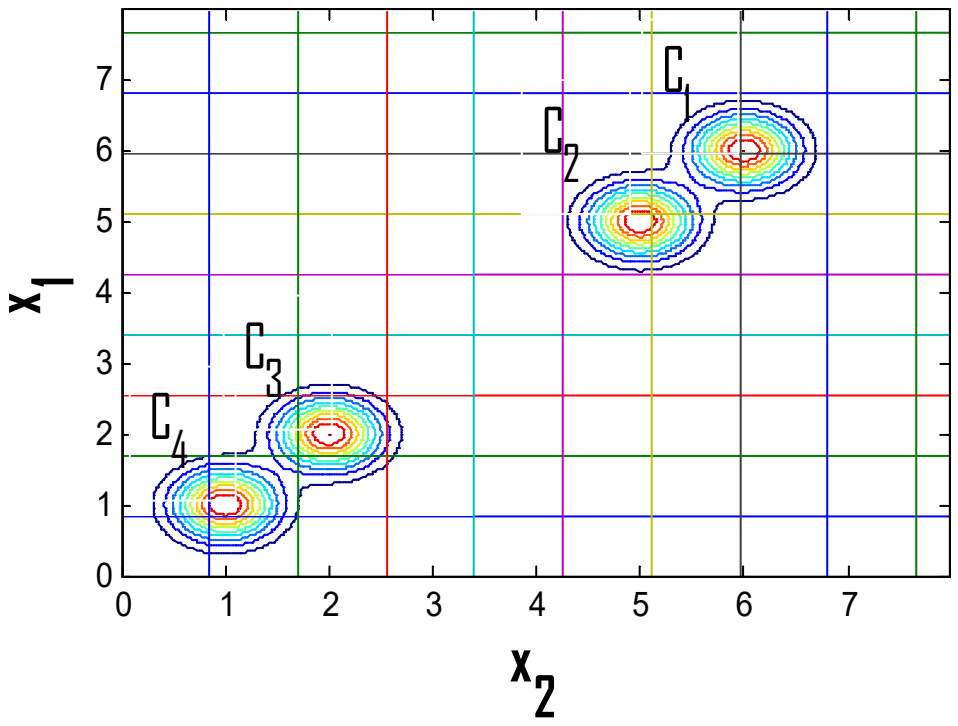

**Figure 4**: 2-dimensional example: initial hyperplanes state, uniform orthogonal grid case

Refer now to Figure 4 which shows a 2-dimensional feature space with 4 classes $C_1$, $C_2$, $C_3$ and $C_4$ depicted by their iso-density contours. In accordance with our proposed model a pool of hyperplanes is initially distributed across the feature space hypercubic domain of interest $\Omega \subset R^d$ with hypercube edge size o (o ≈ 8 in this example) as shown in Figure 4. The initial state of this pool of hyperplanes may be uniformly random, or ordered in e.g. orthogonal gridlines (as we opted herein). In this later case, an approximate guideline which sets the minimal required density of hyperplanes in feature space is that the distance between neighbor parallel hyperplanes be less than the lower bound for the distance between any two classes' modes peaks. The computational complexity of our proposed model is at most quadratic in the dimension d of the feature space. To see this we note that all described operations for a single hyperplane (Equations (4) to (16) above) scale up at most linearly with dimension d, and that the number of hyperplanes itself, needed to create a uniform orthogonal grid such as that of Figure 4, also scales up linearly with d, overall resulting in quadratic complexity. We note that even though each hyperplane is checked upon arrival of a new feature vector, only a small minority is anticipated to require actual update and full calculations execution, namely only those which are sufficiently close to the incoming vector; thus, great part of the computations are skipped most of the time. In fact our simulations with differing dimension feature spaces indicate a complexity scaling up at approximately $O(d^{1.2})$. As we will see later in the sequel, it is possible to exploit only a small subset of the hyperplane pool in order to perform classification; by so doing, complexity during the classification stage may be further greatly reduced. The memory requirement of our proposed model does not depend on the number of input samples (in contrast with offline unsupervised learning schemes such as parametric models or k-Means) and is quadratic in feature space dimension d: the variables needed to be kept for each hyperplane are all of dimension d (or scalar) and the number of hyperplanes scale up linearly with d, resulting in $O(d^2)$. Since the proposed model operates on a sample by sample basis as mentioned above, then continuous learning and adaptation to slowly changing environments are made possible. Our proposed classifier could be used in conjunction with 'human engineered' feature extractors (such as SIFT in the Machine Vision domain), or 'machine learned' representations (such as those associated with Deep NN, with VAE or with GAN as mentioned above); this would result in an end-to-end unsupervised learning pattern recognition machine, eliminating the need for large sets of labeled data.

The proposed model, just as any other unsupervised learning classifier (e.g. k-Means, etc.), generates an output code **y** which results in some arbitrary 'machine assigned' labels set, and which has to be associated with 'human assigned' labels in order to be usefully interpreted by humans. A brief description of the method we have used for this purpose follows: we allocate, after convergence, a bunch of samples along with their 'human assigned' labels for this task (say 100 labeled samples per Class). For every Class we check each hyperplane counting the number of its consistent outputs (say +1) for samples of this current Class and the number of opposite outputs (say -1) for samples not belonging to this Class (a 1-vs-all scheme). We calculate a weighted sum of these 2 numbers and select the highest score hyperplane as this Class discriminator, associating to it that 'human-assigned' label. We have alternatively built a hierarchy tree scheme (e.g. Genus, Species, etc. levels) and similarly picked up the best hyperplanes at each level; both schemes yielded very similar results.

## 3 Simulation Results

To demonstrate the ideas and methods presented herein a Matlab based platform was built. Please refer again to Figure 4. The illustrated 2 pairs ($C_1$, $C_2$ and $C_3$, $C_4$) of 2-variate normally distributed classes, all have equal $\sigma^2\mathbf{I}$ covariance matrix (where $\mathbf{I}$ is the identity matrix), equal priors and shifted means (relative to each other). Inter-means distance and $\sigma^2$ were calibrated, so that the optimal classification error probability was approximately 2 x $10^{-2}$. Rough tuning of the parameters was carried out to ensure fast and reliable convergence, ending up in $\varepsilon$ = 0.0033 $\sigma$, $\Phi$ = 2 $\sigma$, $\alpha$ = 0.04 and $\beta$ = 8 $\sigma$ values; the linear dependence on $\sigma$ of $\varepsilon$, $\Phi$ and $\beta$ is useful and intuitively convenient for problem scaling. Hyperplanes pool initial state was set to uniform orthogonal grid with appropriate distances, as already mentioned above and shown in Figure 4.

Figure 5 shows the final hyperplanes state, after 10000 input samples (~2500 vectors for each of said 4 classes) of the 2-dimensional scenario, for which its initial orthogonal grid state was presented in Figure 4. We notice that many (but not all) hyperplanes have migrated to discriminating states. We see that small groups of hyperplanes, like $\mathbf{hp_3}$ and $\mathbf{hp_4}$, did converge to local low probability density regions, in between classes $C_3$, $C_4$ and $C_1$, $C_2$ respectively.

The internal world representation $\mathbf{y}_t \in R^N$ of such a model can be interpreted as a hierarchical and multi-resolution distributed code: hyperplanes like $\mathbf{hp_1}$ for example could be viewed as discriminating between carnivores and herbivores (no herbivore classes are shown in Figure 4), others like $\mathbf{hp_2}$ may distinguish between canines and felines, still others like $\mathbf{hp_3}$ between cats and lions, and so on. Code redundancy is observed: some of the discriminating valleys are occupied by small groups of neighbor, approximately parallel hyperplanes. Similar hyperplanes final state characteristics as noted above may be expected at higher dimensional feature spaces.

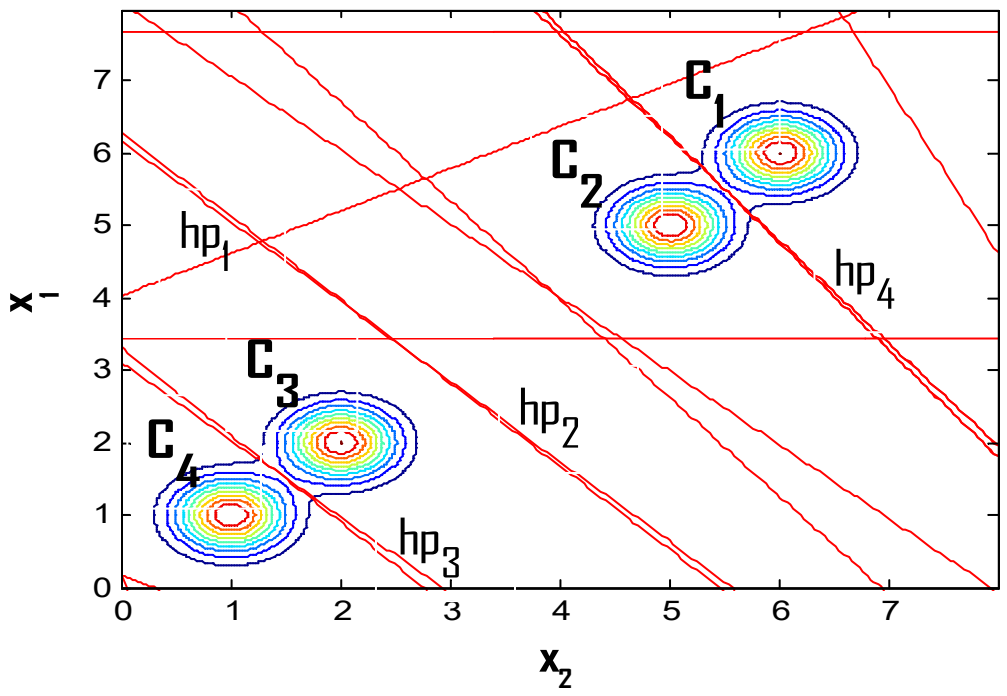

**Figure 5**: 2-dimensional example: final hyperplanes state

Figure 6 shows classification error probability for a similar 2 pair categories scenario as described above, except that dimension now is d = 50. We note $P_{err}$ convergence (in the mean) from 4 x $10^{-1}$ to the optimal 2 x $10^{-2}$ after about 6000 input samples (~1500 for each of 4 classes). Close to optimal $P_{err}$ was expected in this scenario since we chose a setup of equal priors and identical, mean shifted spherically symmetric (normally distributed) conditionals; this symmetry results in a hyperplanar optimal discrimination surface and a collocated hyperplanar valley. $P_{err}$ in Figure 6 shows slight but visible fluctuation around a mean value of 2 x $10^{-2}$. In order to mitigate these fluctuations $\varepsilon$ and $\alpha$, rather than staying fixed during model convergence (as was done in the example of Figure 6) may be made to gradually decrease, either as function of time (sample number) or in some adaptive scheme.

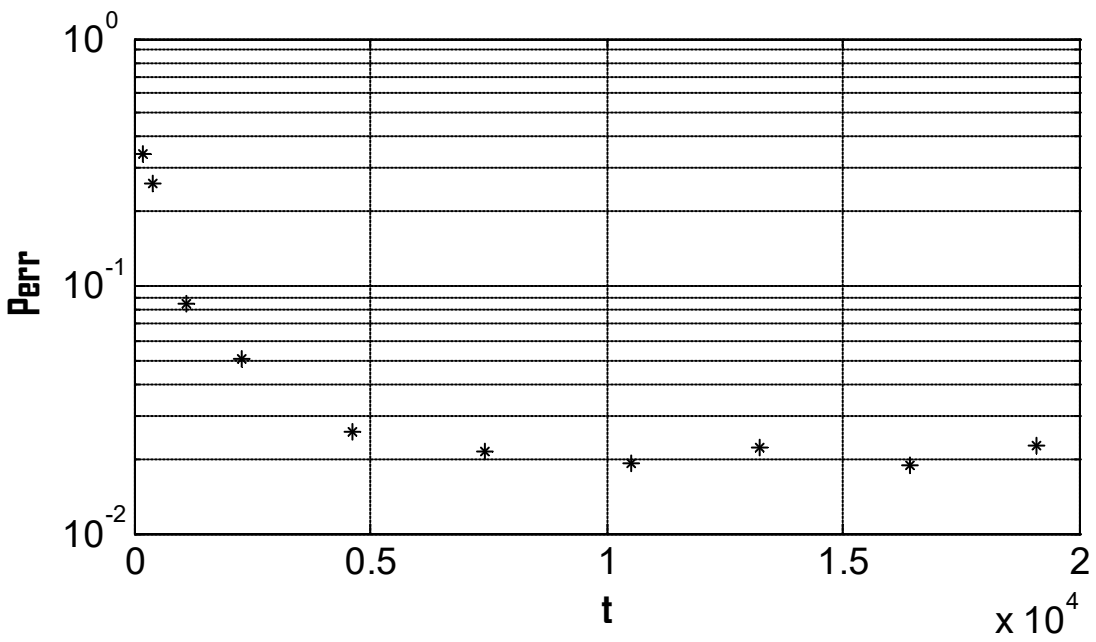

**Figure 6**: 50-dimensional example: Classification Error Probability convergence

To demonstrate performance on a real life application we chose the ImageNet data-set, a challenging popular benchmark containing over 1000 classes of images, with around 1300 labeled samples per class (Deng et al. [7]). Each sample is a full resolution image with average size of 400 x 350 pixels. Our proposed model is generic, and operates in conjunction with any reasonable feature extraction method which yields representations fulfilling our above stated assumptions. We used the (N-1) activation layer of a pre-trained ResNet-50, a variant of which won the ILSVRC 2015 Classification Task (He et al. [8]), with feature space of dimension d = 2048. Inspection and simple analysis of the feature vectors indicate that they may be enclosed by a hyper-cubic domain with edge o = 4. Following this visual inspection we placed 4 hyper-planes per dimension on said initial uniform orthogonal grid reaching a total of 2048 x 4 = 8192 hyperplanes. We left the other hyper-parameters values unchanged, as described above, setting $\sigma = 0.8$. In order to allow easy visualization and comparative analysis of the confusion errors we picked 50 classes from the ImageNet set, containing a mix of fine-grained and coarse-grained classes: 5 species of Sharks and Rays, Cocks and Hens, 11 species of Song Birds, etc. We left out 100 samples per class for testing and used the rest, approximately 1200 samples per class for training. The training set was run during a single epoch; multi-epoch training did not improve the results further. Model parameters $\varepsilon$ and $\alpha$ were kept fixed; their gradual decrease with time did not significantly affect the results. The learning process was relatively fast: about 60000 sample vectors (of d = 2048) in 2.5 hours on an i7, 2.7 GHz PC. We tested classification error performance by estimating, using the fore-mentioned test set, Perr (Top-n, n= 1, 3, 5) which indicates the probability that the actual class of a given sample is not contained in the set of n most probable classes.

We have run this same data set, using same feature representations, with k-Nearest Neighbor, a near optimal popular supervised learning benchmark method. The expected performance gap between kNN and its best-in-class alternative (say SVM) is not big for our purpose. The results of this comparative test are brought in Table 1. We note that our proposed method's Perr (Top-3) exceeds by **merely about 2%** that of the kNN near optimal supervised learning classifier. This excess Perr may be called our unsupervised learning penalty or loss. To probe these results a step further we show in Figure 7 the Confusion Matrices for both our method and kNN; for better readability we show 20 x 20 sub-matrices, but similar behavior is observed in the full matrix. Columns denote 'human assigned' (supposedly ground truth) labels and rows denote our 'machine assigned' labels (numbered 1 to 20 for both).

| **50 Classes Mix** | **Top 5 Perr** | **Top 3 Perr** | **Top 1 Perr** |
|---|---|---|---|
| **kNN** (k=3, Euclidean distance) | 0.024* | **0.043***  | 0.16 |
| **Our Me-thod** | 0.035 | **0.062** | 0.23 |

**Table 1**: Classification Task performance of our proposed method vs. kNN for 50 ImageNet coarse/fine grained classes. The Top-3 and Top-5 kNN figures are approximate calculations

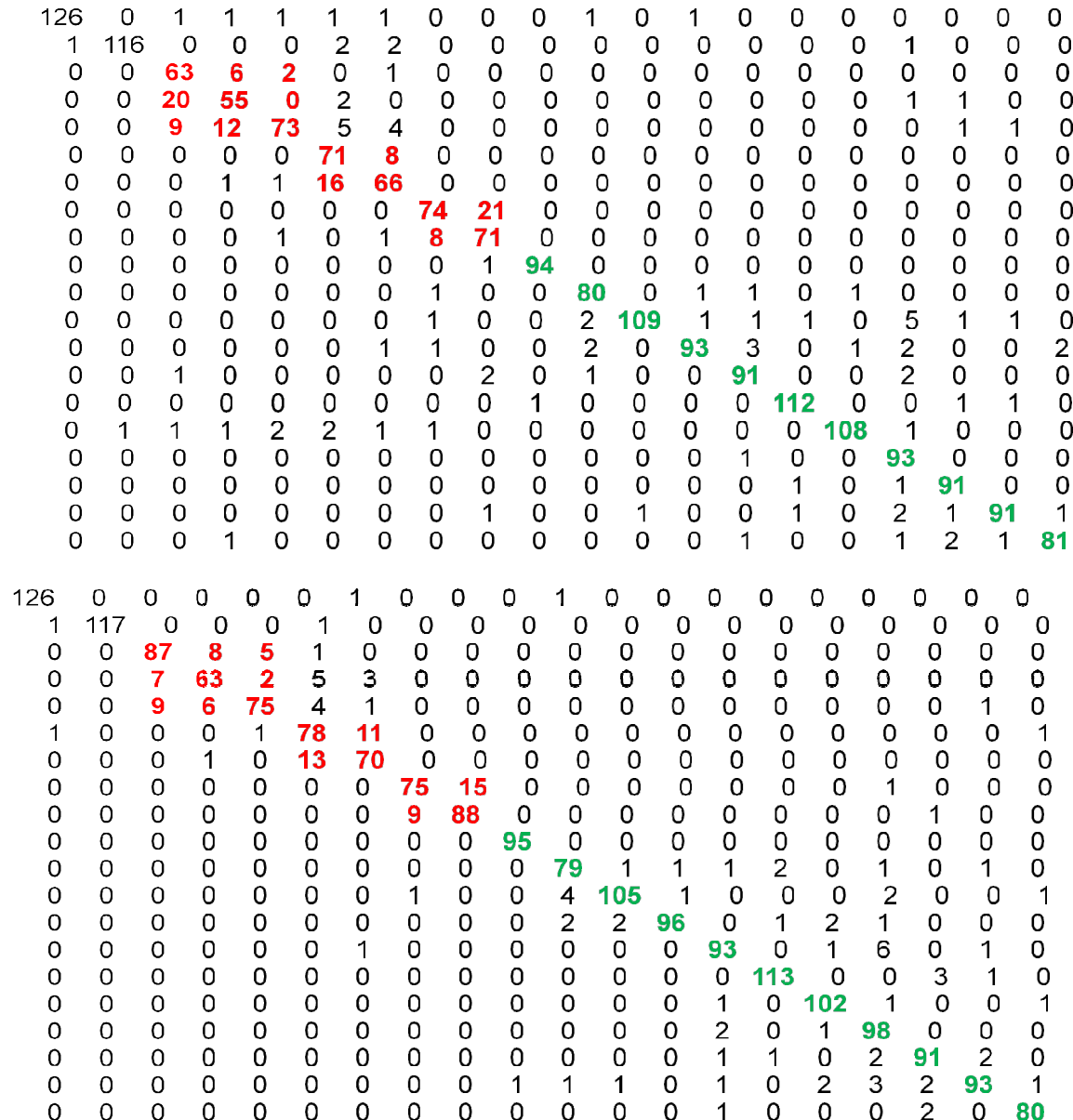

| | | | | | | | | | | | | | | | | | | | |
|---|---|---|---|---|---|---|---|---|---|---|---|---|---|---|---|---|---|---|---|
| 126 | 0 | 1 | 1 | 1 | 1 | 1 | 0 | 0 | 0 | 1 | 0 | 1 | 0 | 0 | 0 | 0 | 0 | 0 | 0 |
| 1 | 116 | 0 | 0 | 0 | 2 | 2 | 0 | 0 | 0 | 0 | 0 | 0 | 0 | 0 | 0 | 1 | 0 | 0 | 0 |
| 0 | 0 | 63 | 6 | 2 | 0 | 1 | 0 | 0 | 0 | 0 | 0 | 0 | 0 | 0 | 0 | 0 | 0 | 0 | 0 |
| 0 | 0 | 20 | 55 | 0 | 2 | 0 | 0 | 0 | 0 | 0 | 0 | 0 | 0 | 0 | 0 | 1 | 1 | 0 | 0 |
| 0 | 0 | 9 | 12 | 73 | 5 | 4 | 0 | 0 | 0 | 0 | 0 | 0 | 0 | 0 | 0 | 0 | 1 | 1 | 0 |
| 0 | 0 | 0 | 0 | 0 | 71 | 8 | 0 | 0 | 0 | 0 | 0 | 0 | 0 | 0 | 0 | 0 | 0 | 0 | 0 |
| 0 | 0 | 0 | 1 | 1 | 16 | 66 | 0 | 0 | 0 | 0 | 0 | 0 | 0 | 0 | 0 | 0 | 0 | 0 | 0 |
| 0 | 0 | 0 | 0 | 0 | 0 | 0 | 74 | 21 | 0 | 0 | 0 | 0 | 0 | 0 | 0 | 0 | 0 | 0 | 0 |
| 0 | 0 | 0 | 0 | 1 | 0 | 1 | 8 | 71 | 0 | 0 | 0 | 0 | 0 | 0 | 0 | 0 | 0 | 0 | 0 |
| 0 | 0 | 0 | 0 | 0 | 0 | 0 | 0 | 1 | 94 | 0 | 0 | 0 | 0 | 0 | 0 | 0 | 0 | 0 | 0 |
| 0 | 0 | 0 | 0 | 0 | 0 | 0 | 1 | 0 | 0 | 80 | 0 | 1 | 1 | 0 | 1 | 0 | 0 | 0 | 0 |
| 0 | 0 | 0 | 0 | 0 | 0 | 0 | 1 | 0 | 0 | 2 | 109 | 1 | 1 | 1 | 0 | 5 | 1 | 1 | 0 |
| 0 | 0 | 0 | 0 | 0 | 0 | 1 | 1 | 0 | 0 | 2 | 0 | 93 | 3 | 0 | 1 | 2 | 0 | 0 | 2 |
| 0 | 0 | 1 | 0 | 0 | 0 | 0 | 0 | 2 | 0 | 1 | 0 | 0 | 91 | 0 | 0 | 2 | 0 | 0 | 0 |
| 0 | 0 | 0 | 0 | 0 | 0 | 0 | 0 | 0 | 1 | 0 | 0 | 0 | 0 | 112 | 0 | 0 | 1 | 1 | 0 |
| 0 | 1 | 1 | 1 | 2 | 2 | 1 | 1 | 0 | 0 | 0 | 0 | 0 | 0 | 0 | 108 | 1 | 0 | 0 | 0 |
| 0 | 0 | 0 | 0 | 0 | 0 | 0 | 0 | 0 | 0 | 0 | 0 | 0 | 1 | 0 | 0 | 93 | 0 | 0 | 0 |
| 0 | 0 | 0 | 0 | 0 | 0 | 0 | 0 | 0 | 0 | 0 | 0 | 0 | 0 | 1 | 0 | 1 | 91 | 0 | 0 |
| 0 | 0 | 0 | 0 | 0 | 0 | 0 | 0 | 1 | 0 | 0 | 1 | 0 | 0 | 1 | 0 | 2 | 1 | 91 | 1 |
| 0 | 0 | 0 | 1 | 0 | 0 | 0 | 0 | 0 | 0 | 0 | 0 | 0 | 1 | 0 | 0 | 1 | 2 | 1 | 81 |
| 126 | 0 | 0 | 0 | 0 | 0 | 1 | 0 | 0 | 0 | 1 | 0 | 0 | 0 | 0 | 0 | 0 | 0 | 0 | 0 |
| 1 | 117 | 0 | 0 | 0 | 1 | 0 | 0 | 0 | 0 | 0 | 0 | 0 | 0 | 0 | 0 | 0 | 0 | 0 | 0 |
| 0 | 0 | 87 | 8 | 5 | 1 | 0 | 0 | 0 | 0 | 0 | 0 | 0 | 0 | 0 | 0 | 0 | 0 | 0 | 0 |
| 0 | 0 | 7 | 63 | 2 | 5 | 3 | 0 | 0 | 0 | 0 | 0 | 0 | 0 | 0 | 0 | 0 | 0 | 0 | 0 |
| 0 | 0 | 9 | 6 | 75 | 4 | 1 | 0 | 0 | 0 | 0 | 0 | 0 | 0 | 0 | 0 | 0 | 0 | 1 | 0 |
| 1 | 0 | 0 | 0 | 1 | 78 | 11 | 0 | 0 | 0 | 0 | 0 | 0 | 0 | 0 | 0 | 0 | 0 | 0 | 1 |
| 0 | 0 | 0 | 1 | 0 | 13 | 70 | 0 | 0 | 0 | 0 | 0 | 0 | 0 | 0 | 0 | 0 | 0 | 0 | 0 |
| 0 | 0 | 0 | 0 | 0 | 0 | 0 | 75 | 15 | 0 | 0 | 0 | 0 | 0 | 0 | 0 | 1 | 0 | 0 | 0 |
| 0 | 0 | 0 | 0 | 0 | 0 | 0 | 9 | 88 | 0 | 0 | 0 | 0 | 0 | 0 | 0 | 0 | 1 | 0 | 0 |
| 0 | 0 | 0 | 0 | 0 | 0 | 0 | 0 | 0 | 95 | 0 | 0 | 0 | 0 | 0 | 0 | 0 | 0 | 0 | 0 |
| 0 | 0 | 0 | 0 | 0 | 0 | 0 | 0 | 0 | 0 | 79 | 1 | 1 | 1 | 2 | 0 | 1 | 0 | 1 | 0 |
| 0 | 0 | 0 | 0 | 0 | 0 | 0 | 1 | 0 | 0 | 4 | 105 | 1 | 0 | 0 | 0 | 2 | 0 | 0 | 1 |
| 0 | 0 | 0 | 0 | 0 | 0 | 0 | 0 | 0 | 0 | 2 | 2 | 96 | 0 | 1 | 2 | 1 | 0 | 0 | 0 |
| 0 | 0 | 0 | 0 | 0 | 0 | 1 | 0 | 0 | 0 | 0 | 0 | 0 | 93 | 0 | 1 | 6 | 0 | 1 | 0 |
| 0 | 0 | 0 | 0 | 0 | 0 | 0 | 0 | 0 | 0 | 0 | 0 | 0 | 0 | 113 | 0 | 0 | 3 | 1 | 0 |
| 0 | 0 | 0 | 0 | 0 | 0 | 0 | 0 | 0 | 0 | 0 | 0 | 0 | 1 | 0 | 102 | 1 | 0 | 0 | 1 |
| 0 | 0 | 0 | 0 | 0 | 0 | 0 | 0 | 0 | 0 | 0 | 0 | 0 | 2 | 0 | 1 | 98 | 0 | 0 | 0 |
| 0 | 0 | 0 | 0 | 0 | 0 | 0 | 0 | 0 | 0 | 0 | 0 | 0 | 1 | 1 | 0 | 2 | 91 | 2 | 0 |
| 0 | 0 | 0 | 0 | 0 | 0 | 0 | 0 | 0 | 1 | 1 | 1 | 0 | 1 | 0 | 2 | 3 | 2 | 93 | 1 |
| 0 | 0 | 0 | 0 | 0 | 0 | 0 | 0 | 0 | 0 | 0 | 0 | 0 | 1 | 0 | 0 | 0 | 2 | 0 | 80 |

**Figure 7**: Our proposed method (top) and kNN (bottom) Confusion (Sub-)Matrices

Adjacent columns (rows) represent similar classes (close animal species in our case), for example columns 3 to 5 carry Shark species, etc. In an errorless scheme the Confusion Matrix would of course be purely diagonal. We notice that both methods exhibit a strikingly similar behavior; regions of greater confusion, such as Sharks (columns 3 to 5), Rays (6, 7) and Cock/Hen (8, 9) have similar levels of confusion; this confusion is mainly due to separability limitations of the ResNet Deep NN furnished feature space. Regions of good error performance (10 to 20, Song Birds) are also remarkably similar in both methods. The small performance gap between kNN and our proposed method is probably mainly due to the fundamental factors mentioned in Section 2 above, namely: non-coincidence between the optimal discriminating hyper-surface and the 'valley' hyper-surface; stochastic fluctuation of our hyperplanes due to finite parameters values, etc.

Given the high dimensionality of the ResNet-50 feature space and the relatively small number of training samples for such a huge space region volume, it is apparent that we have here a case of sparse distribution sampling. One may puzzle how then we could have got such impressive results as presented above; similar good comparative results were also achieved with the MNIST dataset represented by a similarly high dimensionality hand-crafted feature space (benchmarked vs. kNN and SVM, results omitted due to lack of space). A possible explanation is that besides this sparse sampling (sparse in a '1$^{st}$ sense') the feature vectors are also sparse in the sense ('2$^{nd}$ sense') of each containing a relatively large number of zero elements (and few non-zero elements). They are thus confined to low dimensionality manifolds within a much higher dimensional space. When constrained to such manifolds, the space sampling is *effectively* no sparse (in the 1$^{st}$ sense) anymore. We indeed verified that vectors are sparse (in the 2$^{nd}$ sense) by visual inspection of marginal class-conditional densities. An alternative (or additional) reason could be that in spite the insignificant mean number of samples per unit of feature space volume, the share of useful samples which effectively train each hyperplane (approximately $2\Phi/o$), may typically be (as they were for our above shown values of $\Phi$ and o) non-negligible. Evidently more explorative work is required in this area. A similar question regarding learning capability may arise when we notice that the number of model parameters (~ N d ~16e6 in our ImageNet case) is huge relative to the number of training samples (6e4, there); the resolution to this apparent puzzle lies in our view, in the fact that each neuron of our proposed model learns *independently* of each other (as N separate vectors of length d each); this results in d (~ 2e3) parameters being trained by means of 6e4 samples, a reasonable size; this also stands in contrast with 'conventional' multi-layer neural networks, where all neurons are *concurrently* trained (as one long (N d) parameters vector) so that the efficient training set size should be significantly larger than N d.

It is also of importance to evaluate the ImageNet Classification Task performance of other, potentially competing, unsupervised learning models. We pick for that purpose k-Means, possibly the most popular clustering scheme of all. We choose 10 coarse-grained ImageNet classes; these present a simple challenge to our method which yields Perr (Top-1) = 0.015. The k-Means method on the other hand is practically useless at this trivial task as can be seen in Figure 8 which presents k-Means Confusion Matrix. We can readily observe that 'human assigned' Class '1' is split amongst 2 k-Means assigned Classes ('1', '7') and single k-Means Class '6' is assigned to 3 'human assigned' Classes ('6', '7', '8'). This is no surprise; in fact, following these k-Means results we have conducted an exhaustive literature search for works reporting unsupervised learning classification results in conjunction with any feature extractor for ImageNet or other challenging, real life applications: we have found none.

| | | | | | | | | | |
|---|---|---|---|---|---|---|---|---|---|
| **50** | 0 | 0 | 0 | 0 | 0 | 0 | 0 | 0 | 0 |
| 0 | 110 | 0 | 0 | 0 | 0 | 0 | 0 | 0 | 0 |
| 0 | 0 | 88 | 0 | 1 | 0 | 0 | 0 | 0 | 0 |
| 0 | 0 | 0 | 86 | 1 | 0 | 0 | 0 | 0 | 0 |
| 0 | 0 | 2 | 0 | 98 | 1 | 0 | 0 | 2 | 0 |
| 0 | 0 | 0 | 0 | 0 | **100** | **92** | **14** | 0 | 0 |
| **51** | 0 | 0 | 0 | 0 | 0 | 1 | 0 | 0 | 3 |
| 0 | 0 | 0 | 0 | 0 | 0 | 1 | 89 | 0 | 0 |
| 0 | 4 | 1 | 0 | 5 | 0 | 0 | 4 | 95 | 1 |
| 0 | 0 | 1 | 0 | 0 | 0 | 0 | 0 | 0 | 99 |

**Figure 8**: k-Means Confusion Matrix for a simple ImageNet Classification Task for which our proposed method achieves Perr (Top-1) = 0.015

It should be probably evident by now to the reader that this proposed model has a natural neural architecture implementation such as that shown in Figure 9. We note that the resulting neural architecture is feed-forward and 'shallow', consisting merely of a single neural layer.

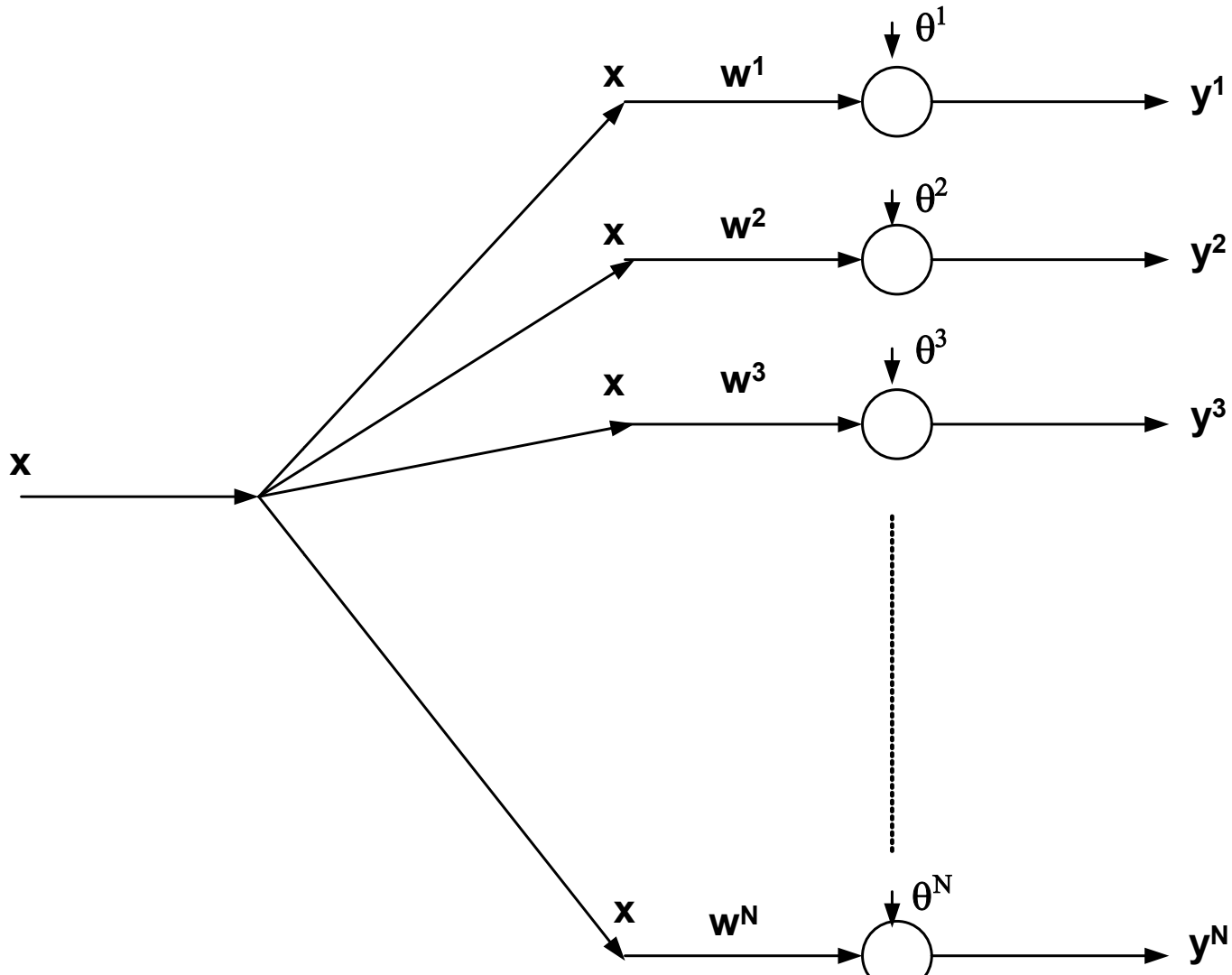

**Figure 9**: Proposed model (shallow) neural architecture

## 4 Concluding Remarks

This work extends the applicability of a linear discriminant surfaces approach to the field of unsupervised learning classifiers. Its main novelty in our view is the *exploitation of the implicit underlying probability density* to train a classifier. We are not aware of any other work which exploits the implicit density for this goal (typically density estimation is carried out). As result of this the proposed method, though stochastic, does not require explicit estimation nor functional form assumption of the probability densities involved. At this initial research stage we have experimentally demonstrated that an unsupervised learning classifier exhibiting low complexity, Hebbian-like local learning rule, online processing, and neural architecture, may achieve competitive classification error performance on a relatively challenging ImageNet Classification Task. Future areas of research may include: test extension to the full ImageNet and other real-life datasets; performance comparison with other supervised learning schemes; feature space sampling sparsity and error performance analysis; convergence analysis; parameters sensitivity analysis; best hyperplane identification analysis; and study of the possibility to extend our model to a near-optimal *supervised* learning variant. Finally, significant progress has been made in recent years in the related fields of unsupervised learning of representations (e.g. VAE [3]) and of supervised learning feature extraction (e.g. GAN [2], DNN Transfer Learning [10]). Integration of either of these 2 types of feature extraction solutions with a classifier model such as ours may bring to life for the first time ever an end-to-end unsupervised learning Pattern Recognition machine with competitive error performance.

For the purpose of reported results replication and model research advance and extension, a Matlab based software package will be provided upon request.

**Acknowledgements.** We are grateful to Meir Feder (Tel Aviv University) for his support and comments and to Yossi Keller (Bar Ilan University) for the provision of ImageNet data.

## References

[1] Duda, R.O., Hart, P.E., Stork, D.G., 'Pattern Classification', 2nd edition, John Wiley & Sons, 2002

[2] Donahue, J., Krahenbuhl, P., Darrell, T., 'Adversarial Feature Learning', arXiv: 1605.09782v6, 2017

[3] Pu, Y., Gan, Z., Henao, R., Yuan, X., Li, C., Stevens, A., Carin, L., 'Variational Autoencoder for Deep Learning of Images, Labels and Captions', NIPS 2016, Barcelona

[4] Ranzato, M.A., Hinton, G.E., 'Modeling Pixels Means and Covariances Using Factorized third Order Boltzmann Machines', IEEE CVPR June 2010

[5] Bishop, C.M., 'Pattern Recognition and Machine Learning', Springer, 2006

[6] Teoh, H.S., 'Formula for vector rotation in arbitrary planes in $R^n$', on the web, unpublished, April 2005

[7] Deng, J., Dong, W., Socher, R., Li, L., Li, K., Fei-Fei, L., 'Imagenet: A Large-Scale Hierarchical Image Database', IEEE CVPR 2009

[8] He, K., Zhang, X., Ren, S. Sun, J., 'Deep Residual Learning for Image Recognition', arXiv: 1512.03385v1, 2015

[9] MacQueen, J. B., 'Some Methods for classification and Analysis of Multivariate Observations', 1967, Proceedings of 5th Berkeley Symposium on Mathematical Statistics and Probability. University of California Press. pp. 281–297

[10] Yosinski, J., Clune, J., Bengio, Y., Lipson, H., 'How transferable are features in deep neural networks?', 2014, Advances in Neural Information Processing Systems 27, NIPS Foundation

[11] Lewis, W., Koontz, G., Fukunaga, K., 'A Non-Parametric Valley-Seeking Technique for Cluster Analysis', 1971, IJCAI'71, Proceedings of the 2nd international joint conference on Artificial intelligence, pp 411-417

[12] Pavlidis, N. G., Hofmeyr, D.P., Tasoulis, S.K., 'Minimum Density Hyperplanes', Journal of Machine Learning Research 17, 2016